\documentclass[conference]{IEEEtran}
\IEEEoverridecommandlockouts
\usepackage{cite}
\usepackage{amsmath,amssymb,amsfonts}
\usepackage{algorithm}
\usepackage{graphicx}
\usepackage{textcomp}
\usepackage{xcolor}
\usepackage{hyperref}

\usepackage{url}
\usepackage{booktabs}
\usepackage{multirow}
\usepackage{threeparttable}

\usepackage{float}
\usepackage{algpseudocode}

\usepackage[whole]{bxcjkjatype} 

\makeatletter
\let\MYcaption\@makecaption
\makeatother

\usepackage{subcaption}

\makeatletter
\let\@makecaption\MYcaption
\makeatother

\def\BibTeX{{\rm B\kern-.05em{\sc i\kern-.025em b}\kern-.08em
    T\kern-.1667em\lower.7ex\hbox{E}\kern-.125emX}}
\begin{document}

\bstctlcite{IEEEexample:BSTcontrol}

\title{
    Feedback is Needed for Retakes: \\
    An Explainable Poor Image Notification Framework \\ 
    for the Visually Impaired
}

\author{
    \IEEEauthorblockN{
        Kazuya Ohata, Shunsuke Kitada, Hitoshi Iyatomi
    }
    \IEEEauthorblockA{
        \textit{Department of Applied Informatics, Graduate School of Science and Engineering, Hosei Univeristy}\\
        Tokyo, Japan \\
        \{kazuya.ohata.2b@stu., shunsuke.kitada.8y@stu., iyatomi@\}hosei.ac.jp
    }
}

\maketitle

\begin{abstract}
    We propose a simple yet effective image captioning framework that can determine the quality of an image and notify the user of the reasons for any flaws in the image.
Our framework first determines the quality of images and then generates captions using only those images that are determined to be of high quality.
The user is notified by the flaws feature to retake if image quality is low, and this cycle is repeated until the input image is deemed to be of high quality.
As a component of the framework, we trained and evaluated a low-quality image detection model that simultaneously learns difficulty in recognizing images and individual flaws, and we demonstrated that our proposal can explain the reasons for flaws with a sufficient score.
We also evaluated a dataset with low-quality images removed by our framework and found improved values for all four common metrics (e.g., BLEU-4, METEOR, ROUGE-L, CIDEr), confirming an improvement in general-purpose image captioning capability.
Our framework would assist the visually impaired, who have difficulty judging image quality.
\end{abstract}


\section{Introduction}

Image captioning technique~\cite{bai2018survey, hossain2019comprehensive} has developed rapidly along with advances in machine learning (ML) technique and has been put to use in practical applications, such as assistive systems~\cite{bai2017smart,ahmed2018image,makav2019new,makav2019smartphone,su2020blindly,ravula2021inverse}, content-based image retrieval~\cite{sindu2019recurrent, piplani2018deepseek}, and agricultural applications~\cite{kumar2019region,marani2021deep,putra2020using}.
As one of the practical applications of this technique, assistance/support for the visually impaired has been attracting attention, and various academic studies~\cite{ahmed2018image, su2020blindly, ravula2021inverse} and industrial application~\cite{taptapsee} have reported their effectiveness.

Assistive systems for the visually impaired could work to verbally inform users about pictures they have taken with their mobile devices.
The system expects the input pictures to be of consistent quality, well focused, and taken under appropriate lighting conditions.
However, there are often difficult cases to analyze in the system due to pictures taken by the visually impaired, such as out-of-focus images, images with suboptimal brightness (e.g.,over- or underexposed brightness), or images that do not show the object that should have been captured in the first place.
Hence, image captioning systems for assisting the visually impaired differ significantly from image input in general captioning tasks.

\begin{figure}[t]
    \centering
    \includegraphics[width=\linewidth]{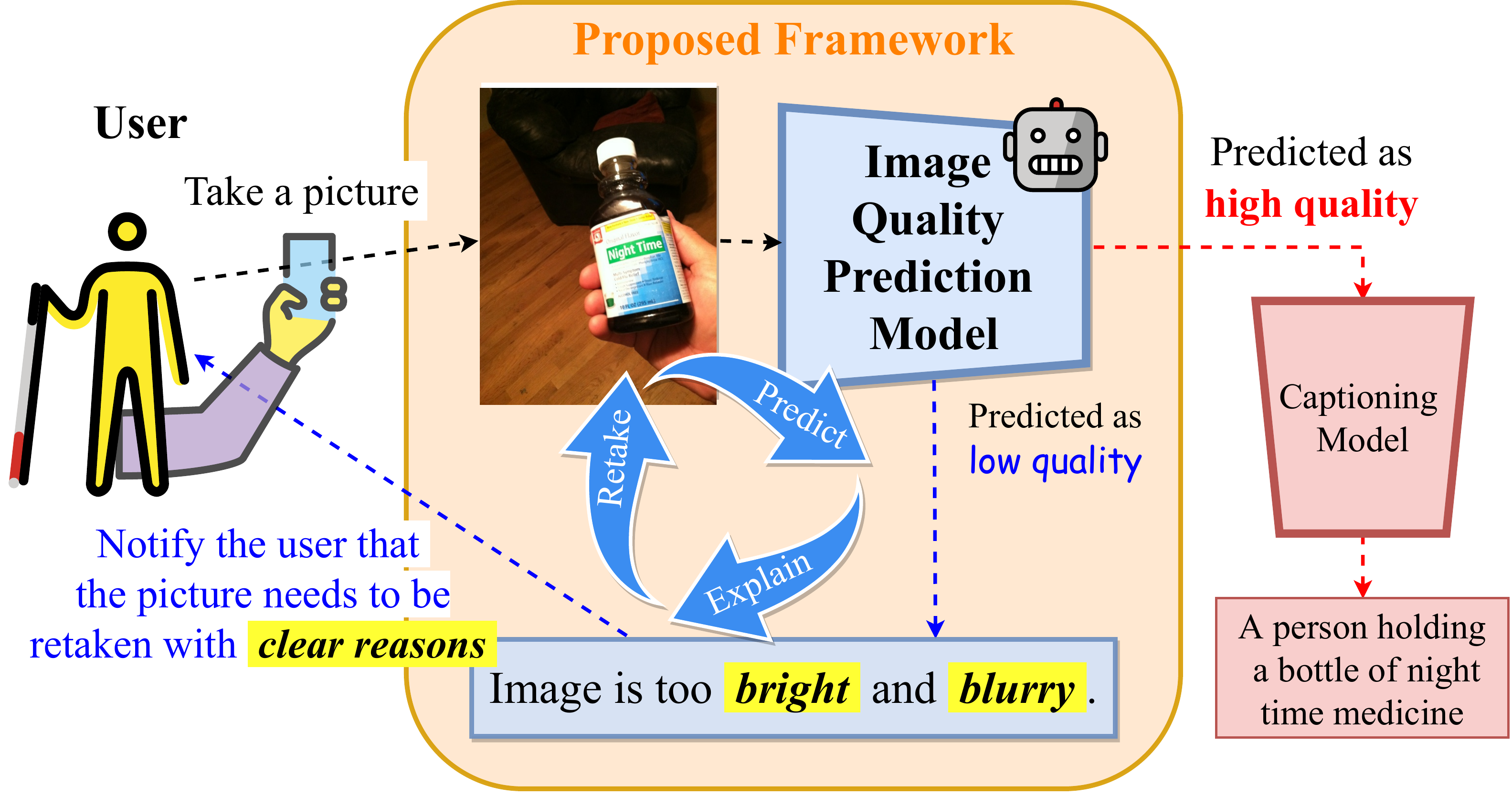}
    \caption{
        Overview of the proposed framework --- PINF ---. 
        To assist the visually impaired, our framework detects images taken under unfavorable conditions and prompts the user to
        retake the picture and tells them why the previous image was deemed poor.
    }
    \label{fig:figure1}
\end{figure}

With the above background, the VizWiz-Captions public dataset~\cite{gurari2020captioning} was released to support the visually impaired, and several studies are using it~\cite{huang2019attention,ahsan2021multi,dognin2022image,wang2022git}.
This dataset contains a variety of images taken under poor conditions by visually impaired individuals.
Additionally, based on the VizWiz-Captions dataset, VizWiz-QualityIssues provided dataset~\cite{chiu2020assessing}, a six-point quantification score on the ``unrecognizable'' and reasons for a total of eight poor image conditions (e.g., blurring).
In recent years, several frameworks~\cite{ye2021novel, dognin2022image} have been proposed improve the detection of objects and characters in images --- using the VizWiz-Captions dataset --- and to generate better image captions.
These frameworks have improved the captioning performance under poor conditions by using models with improved long short-term memory (LSTM)~\cite{hochreiter1997long} or self-attention-based AoANet~\cite{huang2019attention} models.

Although previous studies have reported a certain degree of success in captioning images under poor conditions, we believe there are limits to the improvement.
It is inherently difficult to generate useful captions when low-quality images are input into the system; in such cases, the system typically generates inaccurate captions.
When it is clear from the generated caption that the picture was taken unsuccessfully (e.g., ``black screen'' for a dark picture in which no objects can be recognized), the user can take action (e.g., retaking the picture).
In contrast, an inaccurate caption may cause the user to make a wrong decision.
Since the previous frameworks  have forced to perform prediction even for extremely low quality images, a mechanism needs to notify the user that a correct prediction cannot be made.

In the original paper aided by the VizWiz-QualityIssues dataset~\cite{chiu2020assessing}, an attempt was made to remove these poor images in advance.
Their ResNet152-based~\cite{he2016deep} model achieved F1 = 71.2\% in poor image detection, and it was expected that excluding those low-quality images would improve image captioning performance.
However, no numerical improvement was observed in their experiments.
We believe this result was due to the limited detection ability of those images.

To assist the visually impaired in the use of image captioning technology, we propose a simple yet effective 
framework --- the poor image notification framework (PINF) (as shown in Fig.~\ref{fig:figure1}) --- which detects images taken under unfavorable conditions and prompts the user to retake the picture and tells them why the previous image was deemed poor.
Our PINF applied multi-task learning~\cite{caruana1997multitask}, which can improve prediction performance by learning related tasks simultaneously. 
This allows for more efficient detection of poor images and also provides the user with reasons for undesirable shooting conditions, which is important when building a real-world system.
The proposed framework will give users a better chance of obtaining appropriate captions because they will know what to watch out for and when to re-photograph.

The contributions of this study are summarized as follows:
\begin{itemize}
    \item We proposed PINF, a highly practical support system for the visually impaired based on image captioning technology. Our framework provides the user with a reason for the need for a retake --- a more detailed description of the situation.
    \item Our PINF effectively utilizes multi-task learning and achieves a practical level score of area under the curve (AUC) = 0.924 for poor image detection and mean squared error (MSE) = 0.720 in flaw severity predictions on a scale of 0 to 5 with a single model.
    \item We confirmed that excluding the poor images detected by our framework improves the caption generation performance (approximately improved by 3, 1, 2, and 2 points in BLEU-4~\cite{papineni2002bleu}, METEOR~\cite{banerjee2005meteor}, ROUGE-L~\cite{lin2004rouge}, and CIDEr~\cite{vedantam2015cider}), respectively.
\end{itemize}

    
\section{Proposed Framework}
    We describe our proposed framework --- \textbf{PINF} --- which encourages users to retake images taken under poor conditions and notifies users regarding the reasons for image flaws.

\subsection{Overview of the Proposed Framework}
Fig.~\ref{fig:figure1} shows our PINF for assisting the visually impaired.
The framework notifies the user of the reasons for any flaws in an image, and encourages them when inappropriate to retake the picture, and determines whether the input image is suitable for caption generation.
Our PINF is based on an image quality prediction (IQP) model, which is composed of an image encoder.
The image encoder can use any image recognition model, such as convolutional neural networks (CNNs)~\cite{lecun1998gradient} or vision transformers (ViTs)~\cite{dosovitskiy2020image}.
The final layer consists of a fully coupled layer with 7 outputs, and the following two objectives are estimated simultaneously in this layer: (1) whether the photo was taken under poor conditions and needs to be retaken (1 output) and (2) reasons for image flaws (6 output for each flaws).
PINF repeats quality prediction until the image is judged to be of high quality and passes it to the image captioning model.
This cycle finally allows the user to obtain a reliable caption.

We introduced the idea of the multi-task learning~\cite{caruana1997multitask} technique, where multiple related tasks learn simultaneously, into our framework.
Based on this technique, our proposal is expected to improve the ability of each task by learning the need to retake images and the reasons for flaws.
Our framework has the practical advantage of low execution cost to the user, as it only requires one inference at runtime to generate both results for the user.

\subsection{Image Quality Prediction (IQP) Model}
In the image encoder of the IQP model, we consider three models --- ResNet50~\cite{he2016deep}, EfficientNetB4~\cite{tan2019efficientnet}, and ViT-Base~\cite{dosovitskiy2020image} --- which have been widely successful in the field of computer vision.
The network has $1 + 6 = 7$ outputs for recognition difficulty ``unrecognizable'' and each flaw (framing, blur, dark, bright, obscured, rotation).
We determined the final model based on performance comparison between these three models.
Please note that our PINF is a framework for assisting the visually impaired and is not limited to the settings of both the image encoder and fully connected network.

\subsection{Dataset Containing Flaw Information for Images}
We used the VizWiz-QualityIssues dataset~\cite{chiu2020assessing} of images taken by visually impaired people with labels for image quality.
Each image in the dataset has an ``unrecognizable'' label representing the overall image recognition difficulty and a total of eight image flaw labels (blur, bright, dark, obscured, rotation, framing, others, none), each labeled with a grade from 0 (none) to 5.
We excluded ``others'' labels where the reason for the flaw was not clear, and ``none'' labels where there was no flaw.
Only the training (23,431 samples)  and validation (7,750 samples) data are publicly available; the correct answer labels for the test data are not available.
Thus, we randomly split the validation data in half, using one half as validation data and the other as test data.
The author who presented the dataset~\cite{chiu2020assessing} defined a poor image as one with an ``unrecognizable'' label value of 2 or more, so we set the detection target to 2 or more as well.




\section{Experiments}
\begin{table*}[t]
\centering
\caption{Comparison of detectability of poor condition images for test data}
\label{tab:comparison_detectability}
\begin{tabular}{@{}lrrrrrrrr@{}}
\toprule
\multirow{2}{*}{\begin{tabular}[c]{@{}l@{}}Image encoders \\ in the IQP model\end{tabular}} & \multicolumn{4}{c}{Single-task}                                                                               & \multicolumn{4}{c}{Multi-task}                                                                                \\ \cmidrule(lr){2-5} \cmidrule(lr){6-9}
                                                                                            & \multicolumn{1}{c}{Precision} & \multicolumn{1}{c}{Recall} & \multicolumn{1}{c}{AUC-ROC} & \multicolumn{1}{c}{AUC-PR} & \multicolumn{1}{c}{Precision} & \multicolumn{1}{c}{Recall} & \multicolumn{1}{c}{AUC-ROC} & \multicolumn{1}{c}{AUC-PR} \\ \cmidrule(r){1-1} \cmidrule(lr){2-2} \cmidrule(lr){3-3} \cmidrule(lr){4-4} \cmidrule(lr){5-5} \cmidrule(lr){6-6} \cmidrule(lr){7-7} \cmidrule(lr){8-8} \cmidrule(l){9-9}
ResNet50~\cite{he2016deep}                                                                  & 0.743                         & 0.660             & \textbf{0.919}          & \textbf{0.793}         & 0.694                & 0.704                      & 0.914                   & 0.783                  \\
EfficientNetB4~\cite{tan2019efficientnet}                                                   & 0.664                & 0.752                      & 0.918                   & 0.787                  & 0.696                         & 0.759             & \textbf{0.928}          & \textbf{0.809}         \\
ViT-Base~\cite{dosovitskiy2020image}                                                                                    & 0.658                         & 0.737                      & 0.917                   & 0.777                  & 0.636                         & 0.754                      & 0.909                   & 0.773                  \\ \bottomrule
\end{tabular}
\end{table*}

To validate the effectiveness of our PINF, this section describes two types of experiments: one on the detectability of poor images and the other on the enhancement of captioning performance by excluding such images.

\subsection{Evaluation of Capability to Exclude Poor Images}
We evaluated the ability of our framework to detect low-quality images.
The IQP model is trained with ``unrecognizable'' label and six flaw labels.

We used the following metrics to evaluate the detection of poor images by each model: precision, recall, area under the compute receiver operating characteristic (AUC-ROC), and area under the precision-recall curve (AUC-PR).
To demonstrate the effectiveness of our proposal, we compared our framework with a single-task learning setting in which the ``unrecognizable'' label is trained and evaluated with simultaneous training for six flaws.
In the evaluation of the degree of six flaws in formation, we evaluated the estimation performance with MSE and correlation (Corr.) with the ground truth (GT) label in each flaw.

\subsection{Evaluation of Caption Generation Capability after Exclusion of Poor Images}
We evaluated the caption generation capability with the proposed framework by excluding poor images.
For the caption generation model, we employed ClipCap~\cite{mokady2021clipcap} with Microsoft Common Objects in Context (MSCOCO)~\cite{lin2014microsoft} pre-trained weights, a publicly available state-of-the-art caption generation model.
MSCOCO's pre-trained model was used to measure captioning performance in general settings of image caption generation.
We used the following four common metrics to evaluate image caption generation capability:
\begin{itemize}
    \item \textbf{BLEU-4}~\cite{papineni2002bleu} (4-gram bilingual evaluation understudy). This metric is a precision-based metric, which accounts for precise matching of n-grams in the generated and ground truth references.
    \item \textbf{METEOR}~\cite{banerjee2005meteor} (metric for evaluation of translation with explicit ordering). This metric first creates an alignment between the two sentences by comparing exact tokens, stemmed tokens, and paraphrases.
    \item \textbf{ROUGE-L}~\cite{lin2004rouge} (longest common subsequence version of recall oriented understudy for gisting evaluation). This metric, similar to BLEU, has different n-grams-based versions and computes recall for the generated sentences and the reference sentences.
    \item \textbf{CIDEr}~\cite{vedantam2015cider} (consensus-based image description evaluation). This metric is a human-consensus-based evaluation metric, which was developed specifically for evaluating image captioning methods but has also been used in video description tasks.
\end{itemize}
We adopted these metrics because they have been used in many previous studies and we believe they are sufficiently practical for the evaluation of our framework.

\subsection{Implementation Details}
A regression model was used to predict image quality and MSE was used as the error function.
We trained our model using the Adam~\cite{kingma2014adam} optimizer with a learning rate of 0.00001.
The model was trained using early stopping, which terminates training when the minimum value for the validation data is not updated three consecutive times.
The batch size and number of epochs were set to 128 and 100, respectively.

    
\section{Results}
    \subsection{Detection of Poor Images}\label{sec:comparison_detectability}
\begin{figure}[t]
    \centering
    \begin{minipage}{0.49\linewidth}
        \includegraphics[width=\linewidth]{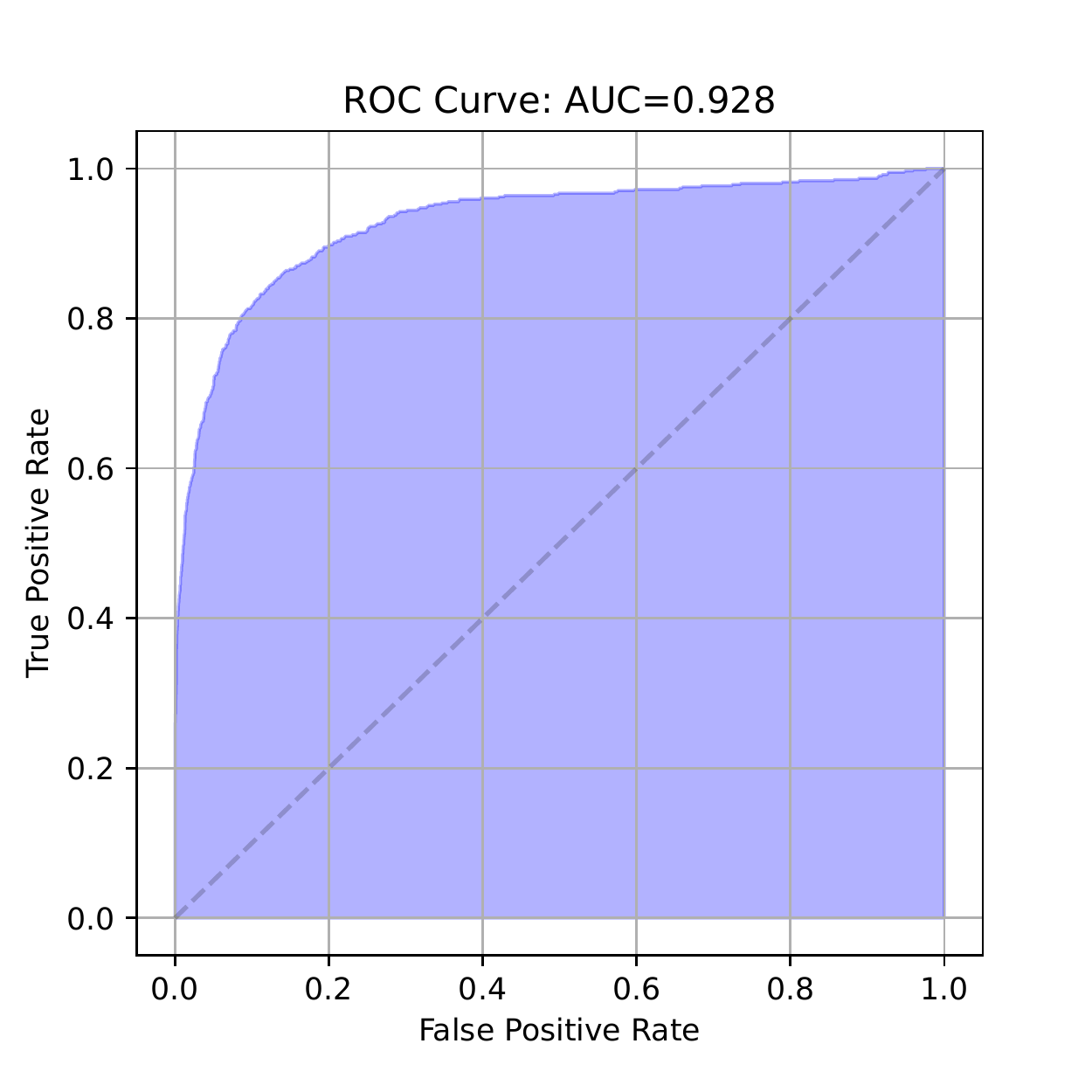}
        \subcaption{AUC-ROC}
    \end{minipage}
    \begin{minipage}{0.49\linewidth}
        \includegraphics[width=\linewidth]{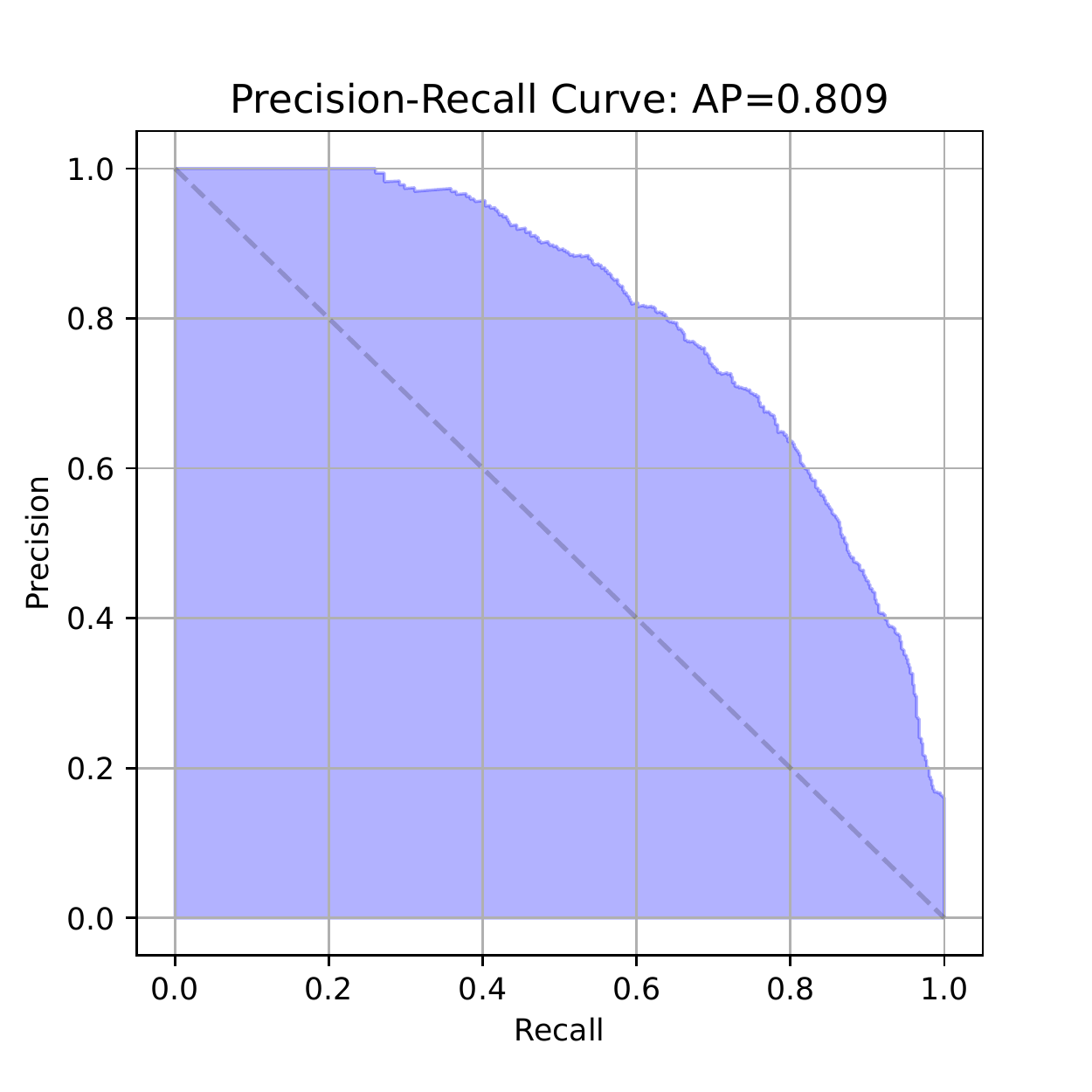}
        \subcaption{AUC-PR}
    \end{minipage}
    \caption{AUC-ROC and AUC-PR for test data based on the proposed framework with the best multi-task EfficientNetB4 encoder.}
    \label{fig:roc_and_pr_test}
\end{figure}

Table~\ref{tab:comparison_detectability} shows a comparison of the detectability of poor images.
The single-task setting uses only the ``unrecognizable'' label, while our multi-task setting uses that label along with six types of flaws information.
The test precision and recall scores were calculated based on the decision threshold when the precision $\times$ recall score reached its maximum in the validation data.
These results show that the performance differences by model and training strategy are not large, but the best scores are provided by the single-task ResNet and multi-task EfficientNet.
Here, we should note that multi-task models have a capability in estimating the degree of flaws simultaneously.
Fig.~\ref{fig:roc_and_pr_test} shows the AUC-ROC and AUC-PR for the test data when trained with multi-task learning using the EfficientNetB4 encoder.



Table~\ref{tab:statistics_and_evaluation} shows the statistics of the flaws and their estimation results. 
The statistics include the mean and S.D. of the ground truth. 
For all flaw information, the MSE score is below the S.D. of each data and the correlation with GT is also at a favorable level.


\begin{table}[t]
\centering
\caption{Statistics of flaw information and estimation performance}
\label{tab:statistics_and_evaluation}
\begin{tabular}{@{}lrrr@{}}
\toprule
\multicolumn{1}{c}{\multirow{2}{*}{}} & \multicolumn{1}{c}{Dataset statistics} & \multicolumn{2}{c}{Evaluation results}                                                                 \\ \cmidrule(r){2-2} \cmidrule(l){3-4}
\multicolumn{1}{c}{}                  & \multicolumn{1}{c}{Mean$\pm$S.D.}      & \multicolumn{1}{c}{MSE} & \multicolumn{1}{c}{\begin{tabular}[c]{@{}c@{}}Corr. \\ with GT\end{tabular}} \\ \cmidrule(r){1-1} \cmidrule(lr){2-2} \cmidrule(lr){3-3} \cmidrule(lr){4-4}
Unrecognizable                        & $0.658 \pm 1.216$                      & $0.502$                 & $0.810$                                                                      \\ \cmidrule(r){1-1} \cmidrule(lr){2-2} \cmidrule(lr){3-3} \cmidrule(lr){4-4}
Framing                               & $1.870 \pm 1.491$                      & $1.436$                 & $0.608$                                                                      \\
Blur                                  & $1.592 \pm 1.701$                      & $1.260$                 & $0.754$                                                                      \\
Dark                                  & $0.304 \pm 0.751$                      & $0.332$                 & $0.638$                                                                      \\
Bright                                & $0.306 \pm 0.739$                      & $0.381$                 & $0.548$                                                                      \\
Obscured                              & $0.200 \pm 0.589$                      & $0.280$                 & $0.443$                                                                      \\
Rotation                              & $0.631 \pm 1.188$                      & $0.848$                 & $0.637$                                                                      \\ \cmidrule(r){1-1} \cmidrule(lr){2-2} \cmidrule(lr){3-3} \cmidrule(lr){4-4}
Average                               & $0.794 \pm 1.096$                      & $0.720$                 & $0.634$                                                                      \\ \bottomrule
\end{tabular}
\end{table}

\subsection{Effects of Caption Generation Capability on the Exclusion of Poor Images}
Table~\ref{tab:vizwiz_vs_qualified_vizwiz} summarizes the caption generation capability of the VizWiz-Captions dataset with and without the exclusion of poor images.
The VizWiz-Captions dataset is the entire test data, and the qualified VizWiz-Captions dataset excluded poor images via our PINF.

By applying PINF, 610 of the 3,875 cases of total test data in the VizWiz-Captions dataset were detected and excluded as poor images, resulting in the Qualified VizWiz-Captions dataset of 3,265 images.
With the poor image exclusion, our PINF with the EfficientNetB4 encoder with multi-task learning performed better than the VizWiz-Captions dataset on all four metrics commonly used in image caption generation tasks.


\begin{table}[t]
\centering
\begin{threeparttable}
\caption{Comparison of caption generation performance between the original VizWiz-Captions dataset and our Qualified VizWiz-Captions dataset}
\label{tab:vizwiz_vs_qualified_vizwiz}
\begin{tabular}{@{}lrr@{}}
\toprule
                                            & \multicolumn{1}{c}{\begin{tabular}[c]{@{}c@{}}Original \\ VizWiz-Captions dataset\end{tabular}} & \multicolumn{1}{c}{\textbf{\begin{tabular}[c]{@{}c@{}}$^\dagger$Qualified \\ VizWiz-Captions dataset\end{tabular}}} \\ \cmidrule(r){1-1} \cmidrule(lr){2-2} \cmidrule(l){3-3}
BLEU-4 $\uparrow$ \cite{papineni2002bleu}   & 54.74                              & \textbf{57.82}                                                                                   \\
METEOR $\uparrow$ \cite{banerjee2005meteor} & 14.82                              & \textbf{15.59}                                                                                            \\
ROUCE-L $\uparrow$ \cite{lin2004rouge}      & 37.32                              & \textbf{39.18}                                                                                   \\
CIDEr $\uparrow$ \cite{vedantam2015cider}   & 27.67                              & \textbf{29.61}                                                                                   \\ \bottomrule
\end{tabular}
\begin{tablenotes}
\item[$\dagger$] Poor images were identified and excluded from the VizWiz-Captions dataset~\cite{gurari2020captioning}.
\end{tablenotes}
\end{threeparttable}
\end{table}


\subsection{Concrete Examples}
\begin{figure*}[t]
    \centering
    \begin{minipage}{0.235\linewidth}
        \includegraphics[width=\linewidth]{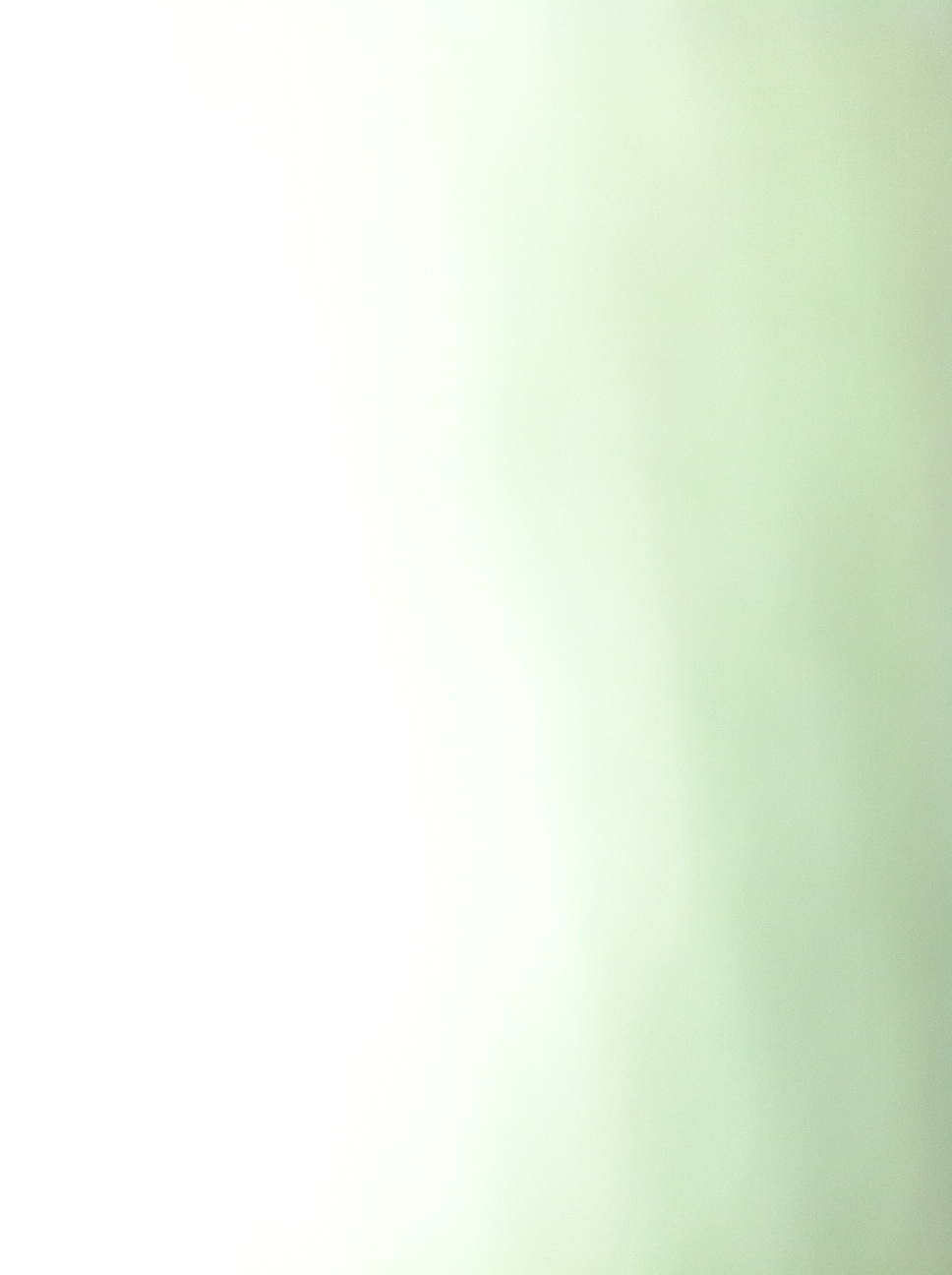}
        \captionsetup{justification=centering}
        \subcaption{\footnotesize GT = 5, Pred. = 4.84 \\ Quality issues are too severe to recognize visual content.}
        \label{fig:correct_01}
    \end{minipage}
    \hspace{0.1cm}
    \begin{minipage}{0.235\linewidth}
        \includegraphics[width=\linewidth]{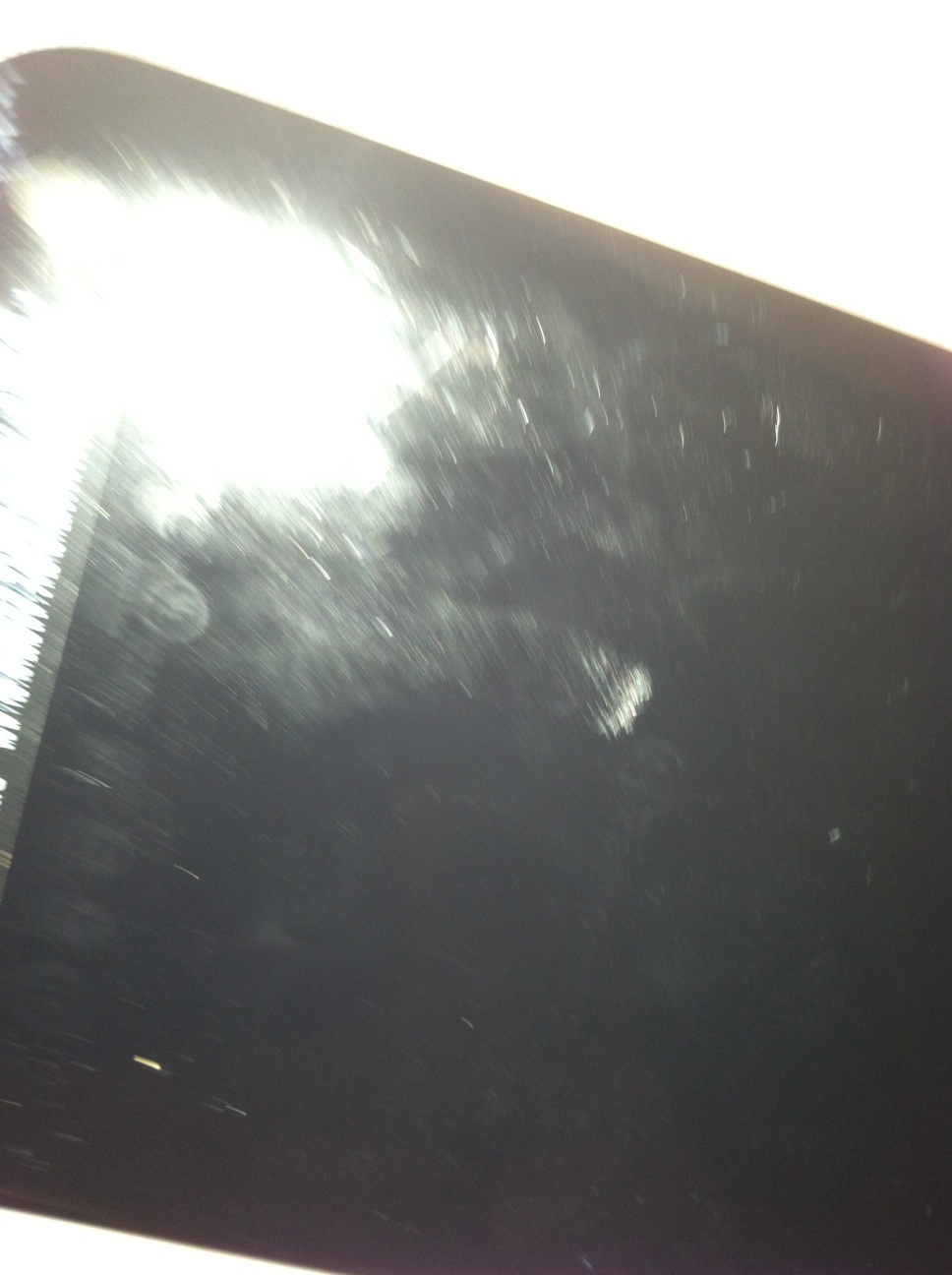}
        \captionsetup{justification=centering}
        \subcaption{\footnotesize GT = 3, Pred. = 2.35 \\ The dark screen of a cell phone is highlighted by white glare.}
        \label{fig:correct_02}
    \end{minipage}
    \hspace{0.1cm}
    \begin{minipage}{0.235\linewidth}
        \includegraphics[width=\linewidth]{figures/jpeg/VizWiz_val_00007507.jpeg}
        \captionsetup{justification=centering}
        \subcaption{\footnotesize GT = 2, Pred. = 2.87 \\ A cardboard looking object with a red and white color pattern on it.}
        \label{fig:correct_03}
    \end{minipage}
    \hspace{0.1cm}
    \begin{minipage}{0.235\linewidth}
        \includegraphics[width=\linewidth]{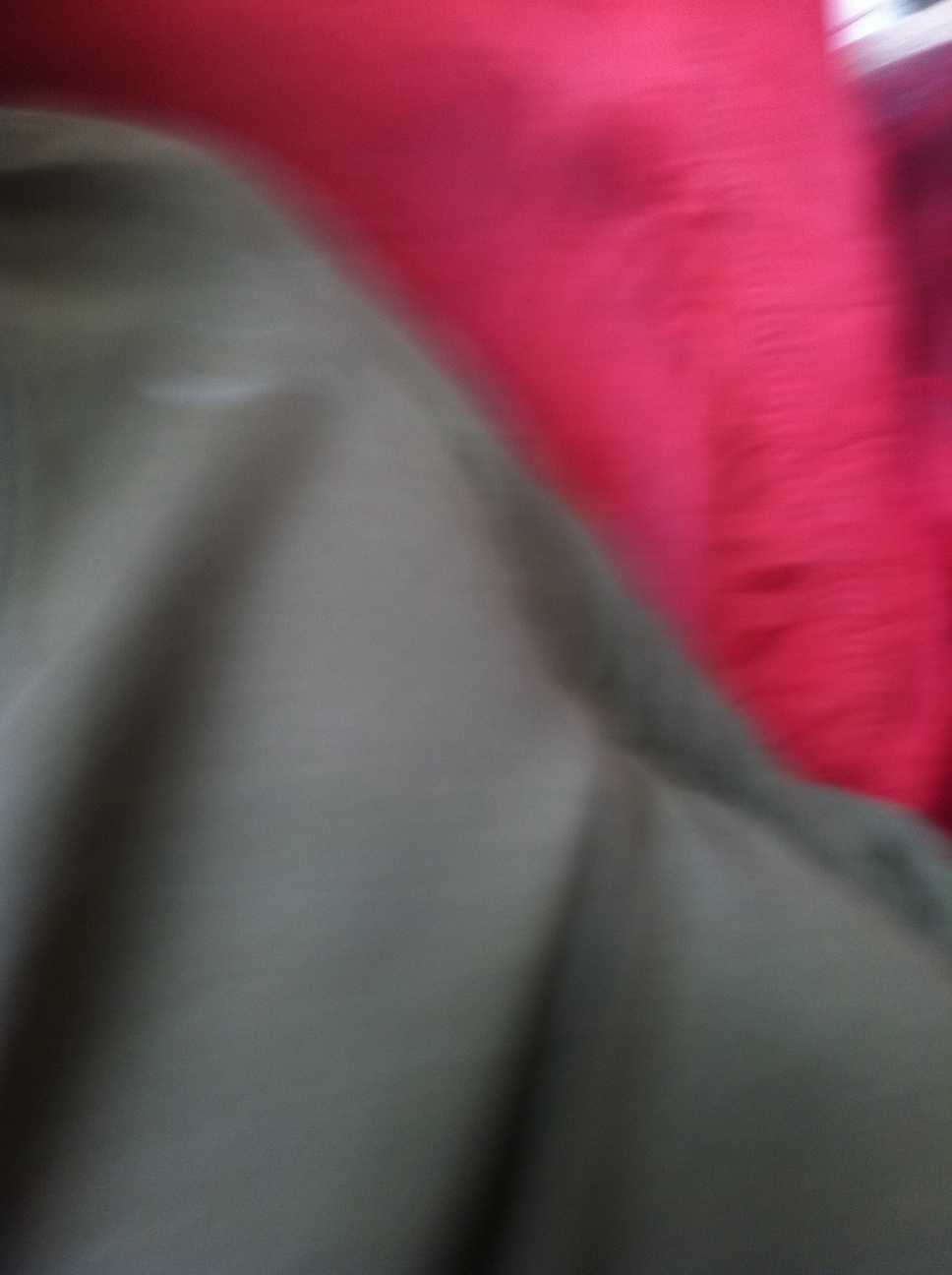}
        \captionsetup{justification=centering}
        \subcaption{\footnotesize GT = 4, Pred. = 3.48 \\ A gray puffy fabric is on top of a red fabric.}
        \label{fig:correct_04}
    \end{minipage}
    \caption{Examples of successfully detected images in the low-quality image exclusion experiment. The ground truth and prediction of the ``unrecognizable'' label as well as the ground truth caption for each image is shown.}
    \label{fig:correct_predictions}
\end{figure*}

\begin{figure}[t]
    \centering
    \begin{minipage}{0.45\linewidth}
        \includegraphics[width=\linewidth]{figures/jpeg/VizWiz_val_00005680.jpeg}
        \captionsetup{justification=centering}
        \subcaption{\footnotesize GT = 1, Pred. = 3.52 \\ A white colored sock on the top of picture and black space on the bottom.}
        \label{fig:incorrect_01}
    \end{minipage}
    \hspace{0.2cm}
    \begin{minipage}{0.45\linewidth}
        \includegraphics[width=\linewidth]{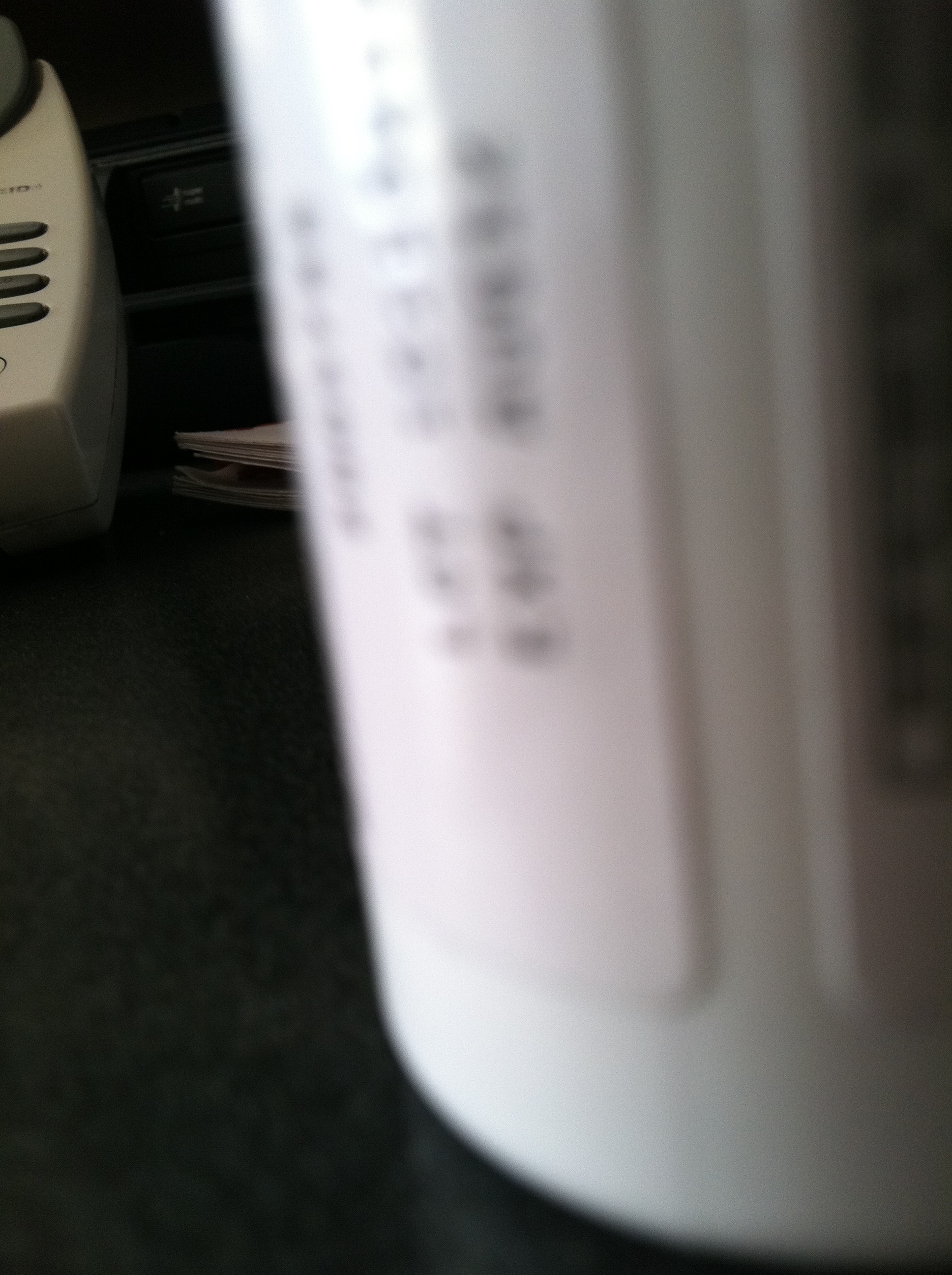}
        \captionsetup{justification=centering}
        \subcaption{\footnotesize GT = 4, Pred. = 1.1 \\ A white object with some blurred lettering is setting against a dark backdrop.}
        \label{fig:incorrect_02}
    \end{minipage}
    \caption{Examples of images that failed to be detected in the low-quality image exclusion experiment. The ground truth and prediction of the unrecognizable, and the ground truth caption for each image is shown.}
    \label{fig:incorrect_predictions}
\end{figure}

Figs.~\ref{fig:correct_predictions} and \ref{fig:incorrect_predictions} show examples of successfully detected poor images failed predictions (over-detection and under-detection) and tables~\ref{tab:correct_predictions} and \ref{tab:incorrect_predictions} show their prediction results and correct scores.
It can be seen that for most items in most examples in both figures, the predicted residuals are kept at a constant level. While the results of the present study are reasonable overall, there are scattered cases in which the high quality indicated by GT (GT = 1) is unconvincing, as in the example in Fig.~\ref{fig:incorrect_01}.



\begin{table}[t]
\centering
\begin{threeparttable}
\caption{Predicted scores on successfully detected poor images}
\label{tab:correct_predictions}
\begin{tabular}{@{}llrrrrrrr@{}}
\toprule
                                           &      & \multicolumn{1}{l}{$^\dagger$Unrec.} & \multicolumn{1}{l}{Frm.} & \multicolumn{1}{l}{Blr.} & \multicolumn{1}{l}{Drk.} & \multicolumn{1}{l}{Brt.} & \multicolumn{1}{l}{Obs.} & \multicolumn{1}{l}{Rot.} \\ \cmidrule(r){1-2} \cmidrule(lr){3-3} \cmidrule(lr){4-4} \cmidrule(lr){5-5} \cmidrule(lr){6-6} \cmidrule(lr){7-7} \cmidrule(lr){8-8} \cmidrule(l){9-9}
\multirow{3}{*}{Fig. \ref{fig:correct_01}} & GT   & 5                          & 3                        & 2                        & 0                        & 5                        & 1                        & 0                        \\
                                           & Pred. & 4.84                       & 1.38                     & 3.34                     & 0.32                     & 2.03                     & 0.78                     & 0.18                     \\
                                           & Diff. & 0.16                       & 1.62                     & 1.34                     & 0.32                     & 2.97                     & 0.22                     & 0.18                     \\ \cmidrule(r){1-2} \cmidrule(lr){3-3} \cmidrule(lr){4-4} \cmidrule(lr){5-5} \cmidrule(lr){6-6} \cmidrule(lr){7-7} \cmidrule(lr){8-8} \cmidrule(l){9-9}
\multirow{3}{*}{Fig. \ref{fig:correct_02}} & GT   & 3                          & 3                        & 3                        & 1                        & 2                        & 0                        & 0                        \\
                                           & Pred. & 2.35                       & 2.56                     & 2.29                     & 0.99                     & 0.8                      & 0.41                     & 0.4                      \\
                                           & Diff.& 0.65                       & 0.44                     & 0.71                     & 0.01                     & 1.2                      & 0.41                     & 0.4                      \\ \cmidrule(r){1-2} \cmidrule(lr){3-3} \cmidrule(lr){4-4} \cmidrule(lr){5-5} \cmidrule(lr){6-6} \cmidrule(lr){7-7} \cmidrule(lr){8-8} \cmidrule(l){9-9}
\multirow{3}{*}{Fig. \ref{fig:correct_03}} & GT   & 2                          & 2                        & 5                        & 0                        & 0                        & 1                        & 0                        \\
                                           & Pred. & 2.87                       & 2.94                     & 4.59                     & 0.11                     & 0.88                     & 0.51                     & 1                        \\
                                           & Diff. & 0.87                       & 0.94                     & 0.41                     & 0.11                     & 0.88                     & 0.49                     & 1                        \\ \cmidrule(r){1-2} \cmidrule(lr){3-3} \cmidrule(lr){4-4} \cmidrule(lr){5-5} \cmidrule(lr){6-6} \cmidrule(lr){7-7} \cmidrule(lr){8-8} \cmidrule(l){9-9}
\multirow{3}{*}{Fig. \ref{fig:correct_04}} & GT   & 4                          & 2                        & 5                        & 0                        & 0                        & 0                        & 0                        \\
                                           & Pred. & 3.48                       & 2.89                     & 3.89                     & 0.28                     & 0.59                     & 1.03                     & -0.01                    \\
                                           & Diff. & 0.52                       & 0.89                     & 1.11                     & 0.28                     & 0.59                     & 1.03                     & 0.01                     \\ \bottomrule
\end{tabular}
\begin{tablenotes}
\item[$\dagger$] Pred., Diff., Unrec., Frm., Blr., Drk., Brt., Obs., and Rot. indicate predicted, difference, unrecognizable, framing, blur, dark, bright, obscured, and rotation, respectively.
\end{tablenotes}
\end{threeparttable}
\end{table}


\begin{table}[t]
\centering
\begin{threeparttable}
\caption{Predicted scores on images that failed detection}
\label{tab:incorrect_predictions}
\begin{tabular}{@{}llrrrrrrr@{}}
\toprule
                                             &      & \multicolumn{1}{l}{$^\dagger$Unrec.} & \multicolumn{1}{l}{Frm.} & \multicolumn{1}{l}{Blr.} & \multicolumn{1}{l}{Drk.} & \multicolumn{1}{l}{Brt.} & \multicolumn{1}{l}{Obs.} & \multicolumn{1}{l}{Rot.} \\ \cmidrule(r){1-2} \cmidrule(lr){3-3} \cmidrule(lr){4-4} \cmidrule(lr){5-5} \cmidrule(lr){6-6} \cmidrule(lr){7-7} \cmidrule(lr){8-8} \cmidrule(l){9-9}
\multirow{3}{*}{Fig. \ref{fig:incorrect_01}} & GT   & 1                          & 2                        & 2                        & 2                        & 0                        & 0                        & 0                        \\
                                             & Pred. & 3.52                       & 2.28                     & 2.83                     & 1.3                      & 0.42                     & 0.64                     & -0.16                    \\
                                             & Diff. & 2.52                       & 0.28                     & 0.83                     & 0.7                      & 0.42                     & 0.64                     & 0.16                     \\ \cmidrule(r){1-2} \cmidrule(lr){3-3} \cmidrule(lr){4-4} \cmidrule(lr){5-5} \cmidrule(lr){6-6} \cmidrule(lr){7-7} \cmidrule(lr){8-8} \cmidrule(l){9-9}
\multirow{3}{*}{Fig. \ref{fig:incorrect_02}} & GT   & 4                          & 3                        & 5                        & 0                        & 0                        & 0                        & 2                        \\
                                             & Pred. & 1.1                        & 3.15                     & 3.97                     & 0.43                     & 0.42                     & 0.31                     & 0.86                     \\
                                             & Diff. & 2.9                        & 0.15                     & 1.03                     & 0.43                     & 0.42                     & 0.31                     & 1.14                     \\ \bottomrule
\end{tabular}
\begin{tablenotes}
\item[$\dagger$] Pred., Diff., Unrec., Frm., Blr., Drk., Brt., Obs., and Rot. indicate predicted, difference, unrecognizable, framing, blur, dark, bright, obscured, and rotation, respectively.
\end{tablenotes}
\end{threeparttable}
\end{table}

\section{Discussion}
    Table~\ref{tab:comparison_detectability} shows that all of the models performed well in detecting poor images.
The multitask learning employed in this study is not necessarily more accurate than single-task learning; it depends on the model.
The single-task ResNet encoder scored as well as the multi-task EfficientNetB4; nevertheless, we believe that the ability to simultaneously estimate the reasons for image flaws in a single model is a major advantage of multi-task learning.
In fact, as shown in Table~\ref{tab:statistics_and_evaluation}, the predicted residuals for scores are significantly smaller than the S.D. of each item in the estimation of the ``unrecognizable'' label and six types of flaws, and positive correlations with GT labels were also observed, suggesting that the accuracy is sufficient for practical use.

Remarkably, Table~\ref{tab:vizwiz_vs_qualified_vizwiz} confirms that the exclusion of poor images by the proposed PINF improved the actual image captioning performance in all the evaluation metrics. 
This highlights the proposed framework's contribution and its potential to support the visually impaired.

In our problem setting, over-detection of poor images is not so bad, although it encourages retakes and increases labor. 
In contrast, lack of detection can cause the user great inconvenience, so it is desirable to minimize the number of cases shown in Fig.~\ref{fig:incorrect_01} as much as possible. 
However, even in this example, our PINF can provide adequate flaw scores, so the user knows that this image may be inappropriate for analysis. 
In this example, the framing and blurring are correctly deemed poor due to multi-task learning reasoning.
    
\section{Conclusion}
    We have proposed a new framework that can effectively assist image captioning technology for the visually impaired.
The proposed PINF can accurately detect poor images that need to be retaken and provide reasons why; moreover, evaluation experiments have confirmed the effectiveness of the PINF framework.
Today, state-of-the-art image captioning technology is capable of highly accurate caption generation for good images.
To support the visually impaired, feedback on retakes is important because it provides the captioning system with only images appropriate for analysis, and the presentation of the reasons for image flaws is an important clue for the user when retaking pictures.


\bibliographystyle{IEEEtran}
\bibliography{references}

\end{document}